\documentclass[conference]{IEEEtran}  

\IEEEoverridecommandlockouts                              

\usepackage{graphicx} 
\usepackage{amsmath} 
\usepackage{tabulary}
\usepackage{soulutf8}

\usepackage{algorithm} 
\usepackage{algpseudocode} 

\usepackage{booktabs}
\usepackage{hyperref}
\usepackage{subfig}

\title{Some essential skills and their combination in an
architecture for a cognitive and interactive robot.}

\author{
\IEEEauthorblockN{Sandra Devin\IEEEauthorrefmark{1}, Gr\'egoire Milliez\IEEEauthorrefmark{1}, Michelangelo Fiore\IEEEauthorrefmark{2}, Aur\'erile Clodic\IEEEauthorrefmark{3} and Rachid Alami\IEEEauthorrefmark{3}}
\IEEEauthorblockA{\IEEEauthorrefmark{1}CNRS, LAAS,
Univ de Toulouse, INP, 7 avenue du colonel Roche, F 31400 Toulouse, France}
\IEEEauthorblockA{\IEEEauthorrefmark{2}CNRS, LAAS,
Univ de Toulouse, INSA, 7 avenue du colonel Roche, F 31400 Toulouse, France}
\IEEEauthorblockA{\IEEEauthorrefmark{3}CNRS, LAAS, 
Univ de Toulouse, LAAS, 7 avenue du colonel Roche, F 31400 Toulouse, France}
}

\begin{document}

\maketitle
\thispagestyle{empty}
\pagestyle{empty}


\begin{abstract}
The topic of joint actions has been deeply studied in the context of Human-Human interaction in order to understand how humans cooperate.
Creating autonomous robots that collaborate with humans is a complex problem, where it is relevant to apply what has been learned in the context of Human-Human interaction. The question is what skills to implement and how to integrate them in order to build a cognitive architecture, allowing a robot to collaborate efficiently and naturally with humans. In this paper, we first list a set of skills that we consider essential for Joint Action, then we analyze the problem from the robot's point of view and discuss how they can be instantiated in human-robot scenarios. Finally, we open the discussion on how to integrate such skills into a cognitive architecture for human-robot collaborative problem solving and task achievement.

\end{abstract}

\section{Introduction}

To interact with its environment, an autonomous robot needs to be able to reason about its surroundings and extract a set of facts that model a semantic representation of the world state. To act on its environment, the robot also needs to decide what actions to take and when to act. Adding humans in the environment requires to enhance the robot with specific skills. Indeed, as humans are social agents, considering them just as moving obstacles or in the same way as a another robot, would lead to dangerous or at least unpleasant interactions for the involved humans. Thus, to estimate the quality of a human-robot collaborative task execution, the robot should not only take into account the task completion but also the comfort and safety of its human partners as well as their acceptability of its actions. A way to increase human comfort when interacting with the robot is to give human-like social skills to it. In this paper, we will present a set of skills needed for Joint Action and that we consider essential and then discuss how they can be implemented and combined into a coherent control architecture.


\section{Requested skills for Joint Actions}

Research in psychology and philosophy have led to a good understanding of human behaviors during joint actions and collaboration. As discussed in \cite{clodic2014key}, the analysis of these studies can help to identify key elements for human-robot joint actions.

The first step when people have to act together is to share a \textit{goal} and the \textit{intention} to achieve it. In \cite{tomasello2005understanding}, Tomassello et al. define a goal as the representation of a desired state and an intention as an action plan chosen in order to reach a goal. Bratman adds that if you have a \textit{shared intention} to perform an action you should agree on the meshing subparts of a \textit{shared plan} \cite{bratman1993shared}. To do so, agents engaged in joint actions need to be able to negotiate the shared plan \cite{pruitt2013negotiation} which requires to have a \textit{common ground} \cite{clark1983common}.

During the execution of the shared plan, agents need to be able to perceive and represent the other agents involved in the task. In \cite{sebanz2006joint}, Sebanz et al. present three necessary skills concerning the perception and representation of the others during joint actions:
\begin{itemize}
\item \textit{Joint attention:} the ability to direct the attention of a partner on a common reference in order to put an element of the interaction (e.g. an entity, a task) in focus.
\item \textit{Action observation:} agents need to be able to recognize others' actions and to understand their effects on the environment and the task. Concerning this, Pacherie discusses two types of predictions in \cite{pacherie2012phenomenology}: \textit{goal-to-action} and \textit{action-to-goal} prediction.
\item \textit{Co-representation:} it is also necessary for an agent to have a representation of others abilities. Knowing what others know, their goals or their capacities, allows to better predict and understand their actions. In other words, agents need to have a \textit{Theory of mind} \cite{baron1985does}. 
\end{itemize}

Finally, during a joint action, agents need to coordinate with each other. In \cite{knoblich20113}, Knoblich et al. define two types of coordination:
\begin{itemize}
\item \textit{Emergent coordination:} it is non voluntary coordination as entrainment (for example, two people in rocking chairs will synchronize their movements) or coordination coming from affordances \cite{gibson1977theory} and from access to resources.
\item \textit{Planned coordination:} when people voluntary change their behavior in order to coordinate. They can do it by adding what Vesper et al. called \textit{coordination smoother} in \cite{vesper2010minimal} or by using verbal or non-verbal communication.
\end{itemize}


\section{From the robot's point of view}

To have an intuitive, natural and fluid interaction with its human partners a robotic system needs to integrate the skills described in the previous section. However, these skills need to be adapted, since a robot may have different capacities than humans, and since humans' behaviors may be different when interacting with a robot, compared to when interacting with another human.

\subsection{Building and maintaining common ground}

The first skill needed by the robot to interact properly with a human is to provide means to understand each other in the situated context of the joint activity. The robot and the human need to share a common ground, meaning that they can identify, in their own world representation, the actions and objects referred to by their counterpart. 
Robotics systems rely on sensors to recognize and to localize entities (humans, robots or objects) in order to build a world state. These sensors produce coordinates to position the objects according to a given frame. For example, a stereo camera with an object recognition software can tell that a mug is in a given position $x$, $y$, $z$ with an orientation of $\theta$, $\phi$, $\psi$.
Humans, instead, use relations between objects to describe their position. Therefore, to indicate the location of a mug, the human would say that it is on the kitchen table, without giving its coordinates or orientation.
To understand the human references and to generate understandable utterances, the robot needs therefore to build a semantic representation of the world, based on the geometric data it collects from sensors, as done in \cite{lemaignan2012grounding}.

We have developed a module based on spatial and temporal reasoning to generate "facts" about the current state of the world \cite{milliez2014framework}. A fact is a property which describes the current state of an entity (e.g. $MUG$ $isNextTo$ $BOTTLE$, $MUG$ $isFull$ $TRUE$). This framework generates facts related to the entities' position and facts about affordances to know, for instance, what is visible or reachable to each agent (human and robot). 
It also generates facts about agent postures to know if an agent is pointing toward an object or where an agent is looking. When the robot tries to understand the human, it should also use these data to improve the information grounding process.

In addition to the world state (which can be considered as the robot's belief state), our situation assessment module maintains a separate and consistent belief state for each human, updated by estimating what the human is able to perceive of the world state. This belief state is an estimation of the list of facts that the agent believes to be true.
In psychology this ability is called conceptual perspective taking.
In \cite{ferreira2015} we have used this framework along with a dialog system to implement a situated dialog and consequently to improve dialog in term of efficiency and success rate.

\subsection{Joint goal establishment}

Once the common ground is established and maintained, the robot needs to share a goal with its human partners. This goal can be imposed by a human (direct order) but the robot should be able to proactively propose its help when needed. To do so, the robot needs to be able to infer high-level goals by observing and reasoning on its human partners' activities. This process is called plan recognition or, when a bigger focus is put on Human-Robot Interaction (HRI) aspects, intention recognition. There is a rich literature on plan recognition, using approaches such as classical planning \cite{ramirez2009plan}, probabilistic planning \cite{bui2003general} or logic-based techniques \cite{singla2011abductive}. In the context of intention recognition work such as \cite{breazeal2009embodied}, and \cite{baker2014modeling} introduced theory of mind aspects in plan recognition.
In our system we provide an intention recognition system, using as inference model a Bayesian Network, which links contextual information, intentions, actions, and low-level observations. A key aspect of the system is using Markov Decision Processes (MDPs) to link intentions to human actions. Each MDP represents an intention, and is able to compute what is the action, in a given situation, that a rational agent would execute to achieve the associated goal.
To properly infer a human agent's intention, it is important to consider his current estimated mental belief in the recognition process, as the actions carried out by the human may be consistent with his intention in his mental representation of the world but not in the actual world state. Our system is able to track and maintain, as discussed earlier, a belief state for each human, which is used as current state for the MDPs when computing the action rewards. 

\subsection{Elaborating and negotiating shared plans}

After establishing a joint goal, in order to achieve it, the robot should be able to create, and negotiate a shared plan \cite{grosz1988plans} with its human partners. To do so, there are several possibilities. The human can choose a specific plan, which the robot needs to be able to understand. Otherwise the robot can compute a plan on its own, which may be negotiated with the human.

\subsubsection{Shared plans elaboration}

In our system, the robot can generate plans using a high level task planner, called Human-Aware Task Planner (HATP) \cite{lallement2014hatp}, which is able to compute a shared plan for a given goal in a given context. HATP can plan actions not only for the robot, but also for its human partners. The computed plan takes into account social rules (such as effort sharing) and also the knowledge of the other agents. For example, in some applications and contexts, the system could try to teach new ways of solving a problem to the human. In that case, the robot will favor plans where the human has to perform new tasks in order to make him learn by experience. The system could also choose that the efficiency is more important than learning in a given situation, in which case HATP will generate a plan to minimize the number of unknown tasks to be performed by the human \cite{Milliez16}.

\subsubsection{Sharing and negotiation}

Once the plan found, it needs to be shared/negotiated with the collaborator in order to be accepted by him.
When dealing with simple plans, infants can cooperate without language (using shared attention, intention recognition and social cues). In situations requiring more complex ones, language is the preferred modality \cite{Warneken2006,Warneken2007}.

There are two possible situations studied in HRI. First, the human has a plan and needs to share it with the robot. Some research in robotics studied how a system could acquire knowledge on plan decomposition from a user \cite{Mohseni2015} and how dialog can be used to teach new plans to the robot and to modify these plans \cite{Petit2012}.
When the robot is the one who has to share its generated plan with the human, to ensure his acceptability and comfort, it should allow the user to contest the plan and to formulate requests on the plan.
To do so we have devised a module able to present the plan higher-level tasks and to ask for the user's approval \cite{Milliez16}. The user can also refuse the plan and ask to perform (or not) specific tasks, which would result in a new plan generation from the robot, trying to take into account the user preference concerning task allocation.

\subsection{Executing shared plans and reacting to humans}

\subsubsection{Maintaining and executing shared plans}

When both agents agree on a plan, the execution process may start. During this execution, the robot needs to coordinate with its human partner. This must be achieved both at task planning level, by synchronizing the robot's plan with the human's, and at a lower level, by adapting the robot's action to the human. In the literature, executives like Pike \cite{karpas2015robust} and Chaski \cite{shah2011improved}, explicitly model humans in their plans and allow the robot to adapt its behavior to their actions. In our system, as explained in \cite{fioreiser2014}, we are able, using monitoring and task preconditions, to perform plan-level coordination. The robot executive decides when to execute the robot actions and assure, at each time, that the plan is still feasible. In case of an unexpected behavior of a human, leading to a plan that is non-valid anymore, the robot is able to quickly find a new plan and adapt to the new situation.

Moreover, the robot should take the human knowledge into account when managing the shared plan. For example, during the execution, it may happen that a human might not know how to perform a task.
In this case, the robot should be able to guide the user to perform the requested task. To make this possible and to adapt the explanation to the human's knowledge on the task to perform, we have created a component which models and maintains the human's knowledge on each task. When a human has to perform a task, the system will check if he knows how to perform it, and adapt its behavior to this information by explaining or not the task \cite{Milliez16}.

The robot executive also models and maintains humans mental states concerning goals, plans and actions. For example, at each time, the robot estimates if its partners are aware of the current plan and if they know which actions have been performed and which ones remain to be done. Then, the robot uses these mental states to detect when a divergent belief can affect the smooth execution of the shared plan (for example if a human does not know that he has to perform an action) and communicates the needed information to inform the human and hence, correct his belief \cite{Devin16}.

\subsubsection{Actions recognition, execution and coordination}

At a lower level, the robot needs to correctly interpret human signals and actions, perform its action in a legible, safe and acceptable way, and, when an action is a joint action (e.g. handover), coordinate its execution with its human partner.

In order to perform actions, in our system, the robot is equipped with a human-aware geometric task and motion planner \cite{sisbot2012human} (both for navigation or manipulation tasks). It also has real time control algorithms to react quickly and stop its movements. As an example, when both the human and itself are trying to execute manipulation operations in the same workspace, the robot would wait for the human to achieve his action to prevent any dangerous situation for the human.

To recognize human actions, the robot needs to observe human movements and infer what action is being performed. An interesting idea is using the robot's own internal models in order to recognize actions and predict user intents, as shown by the \textit{HAMMER} system in \cite{demiris2007prediction}. In our system we perform this process with geometrical reasoning, linking the position and movement of human joints to points in areas in the environments, like objects and rooms. 

When the action is a joint action, like an handover, our system uses Partially Observed Markov Decision Processes (POMDP) to achieve action level coordination and estimate, from observations, the human's engagement level in the action, and to react accordingly \cite{fioreiser2014}. This mechanism was applied in \cite{fiore2015adaptive} to guide a human to a destination in a socially acceptable way, for example by adapting the robot's speed to the human's needs. Our robot is also able to produce proactive behavior when executing joint actions such as suggesting a solution by extending its arm when it needs to perform an handover \cite{pandey2013towards}.

An other way to coordinate is non-verbal communication. In \cite{breazeal2005effects}, Breazeal et al. demonstrate that \textit{explicit} (expressive social cues) as well as \textit{implicit} (communication of implicit mental-state through non-verbal behaviours) communication is useful and necessary if we want a fluent human-robot interaction. The robot needs to be able both to detect and interpret the non-verbal cues given by its partners and to produce non-verbal signals in an understandable and pertinent way. However, these signals need to be adapted to the physiognomy of the robot: for example, for a robot with a head and without eyes, it has been established that gaze signals can be replaced by the orientation of the head \cite{boucher2012reach}.


\section{An Architecture for a cognitive and interactive robot}
\label{c}

We discuss here below how the skills described above can be combined in key elements of a cognitive architecture for a collaborative robot. Pacherie describes a three-level architecture used to monitor human-human joint actions \cite{pacherie2012phenomenology}. It consists of a \textit{Shared Distal level} which deals with joint intention issues (commitment, creation and monitoring of a shared plan), a \textit{Shared Proximal level} which deals with the execution of the plan actions coming from the Distal level at a high level, and a \textit{Coupled Motor intention} which ensures the coordination in space and time during the action execution.

\begin{figure}[!t]
        \centering
          \includegraphics[width=0.4\textwidth]{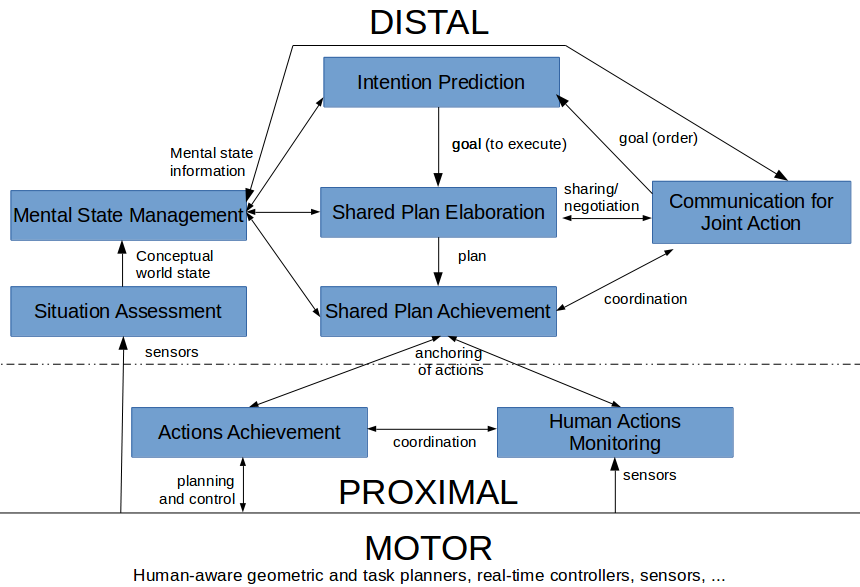}
    \caption{A cognitive architecture for Human-Robot Interaction.}
       \label{fig:archi}
\end{figure}

In robotics, Alami et al. \cite{alami1998architecture} drew an architecture with three similar levels. Then, this architecture has been developed in the context of Human-robot interaction \cite{alami2011robot,alami2013human, fioreiser2014} integrating more and more human-aware skills. Starting from these architectures we tried to design a cognitive system for human-robot interaction. This architecture (Figure \ref{fig:archi}) is intended to equip the robot with all the skills described in the previous section. 

First, the architecture would require a \textbf{Situation Assessment} component to understand the environment in which the interaction takes place. To do so, this component needs to collect data from sensors and to perform spatial and temporal reasoning to describe the environment in term of predicates and properties. These properties should describe the situation of each entities in term of relational position (BOB isIn KITCHEN), and state property (MUG isEmpty FALSE). It should also describe the situation of agents in term of posture (BOB isPointing MUG), affordance (BOB canReach MUG), and motion or actions (BOB isMovingToward MUG).

We previously described how conceptual perspective taking is important to understand humans' utterance or assess humans' intentions. To build and maintain a separate and consistent mental state for each agent, a \textbf{Mental State Management} component is needed. Each Mental state consists of a set of facts related to the environment, like the one generated by the Situation Assessment component, but also of facts concerning what the agent knows about other agents' knowledge on the goal, the plan and tasks included in the plan and about the agents' capacities.

The architecture also requires a component dealing with \textbf{Communication for Joint Actions}. Based on the information concerning the agents' mental states, this component should provide situated dialogue and perspective taking skills when discussing with humans or sharing/negotiating a plan. This component should also allow the robot to produce human-like non-verbal signals and to correctly recognize and interpret the signals of its human partners.

The architecture should also involve a component to determine the robot's current goal. This is the \textbf{Intention Prediction} component. First, this component should be able to recognize  humans' intentions, based on the estimation of their mental states and communicative feedback. Then it should estimate if the robot can and has to help humans and choose the most appropriate goal to execute. During the execution of the chosen goal, this component should estimate if the humans involved into the goal are still (or not) committed to it.

Once a goal is chosen, a \textbf{Shared Plan Elaboration} component would allow the robot to agree on a shared plan with its partners. Therefore, this component should contain a high level task planner, like HATP, capable of computing human-aware plans considering social costs and other agents mental states. This component should also allow the robot to share/negotiate a plan with humans.

Finally, the architecture should contains a component, that we call \textbf{Shared Plan Achievement}, which ensures the smooth execution of the shared plan. This component should consider other agents' mental states to coordinate with them and communicate information when needed. It will send a request to the \textbf{Action Achievement} component to perform robot actions and gather information from \textbf{Human Actions Monitoring} which will interpret humans movements and recognize humans actions. The Action Achievement component should execute the robot's actions in a legible, safe and acceptable way and, when those actions are joint actions (e.g. handover), it should ensure the proper coordination with the human partners.


\section{Conclusion}

In this paper, we briefly reviewed a set of skills that we consider as essential for joint action. Then we discussed how to apply these skills to HRI and how to integrate them into a cognitive architecture. However, in the current robotics context, we are still far from a real cognitive architecture able to perform fluid and friendly Human-Robot joint actions. Each of this skills can be more or less sophisticated and each them is a topic of research by itself.
The next step should be to design and implement, with the robotic community, a complete cognitive architecture inspired from social sciences and neuroscience but adapted to the capacities of a robot and to the behavior of humans, when they interact with robots. 

\addtolength{\textheight}{-11cm}


%
\bibliographystyle{ieeetr}
\bibliography{intention.bib,planning.bib,hrteams.bib,sandra.bib,plan_recognition.bib}





\end{document}